\RequirePackage{fix-cm}
\documentclass[twocolumn]{svjour3}          
\smartqed  
\usepackage{graphicx}
\usepackage{times}
\usepackage{epsfig}
\usepackage{epstopdf}
\usepackage{amsmath}
\usepackage{amssymb}
\usepackage{subfigure}
\usepackage{caption}
\usepackage{tabularx}
\usepackage[table]{xcolor}
 \usepackage[normalem]{ulem}
\usepackage{algorithm}
\usepackage[noend]{algpseudocode}
\usepackage{hyperref}

\usepackage[section]{placeins}
\makeatletter
\def\BState{\State\hskip-\ALG@thistlm}
\makeatother
\definecolor{myGreen}{HTML}{33FF00}
\definecolor{myRed}{HTML}{FF3030}
\definecolor{myGrey}{HTML}{AA5555}
\definecolor{myWhite}{HTML}{FFFFFF}
\definecolor{maroon}{cmyk}{0,0.87,0.68,0.32}
\definecolor{petr}{HTML}{5555FF}
\definecolor{josef}{HTML}{FF3030}

\journalname{xxx}
\begin{document}

\title{AdderNet and Its Minimalist Hardware Design for Energy-Efficient Artificial Intelligence\thanks{Chunjing Xu and Dacheng Tao supervised the project. Yunhe Wang and Mingqiang Huang proposed and designed the experiment, Kai Han, Hanting Chen and Yunhe Wang performed the AdderNet algorithm verification. Mingqiang Huang and Wei Zhang performed the hardware design. Mingqiang Huang, Yunhe Wang, and Kai Han analyzed the data and co-wrote the manuscript. All the authors discussed the results and commented on the manuscript. ‡These authors contributed equally: Yunhe Wang, Mingqiang Huang. *E-mail: dacheng.tao@sydney.edu.au; yunhe.wang@huawei.com; mq.huang2@siat.ac.cn}
}


\author{ Yunhe Wang${}^{1}$${}^{\mathrm{\ddagger }}$, Mingqiang Huang${}^{2}$${}^{\mathrm{\ddagger }}$, Kai Han${}^{1}$, Hanting Chen${}^{1}$, Wei Zhang${}^{2}$,\\Chunjing Xu${}^{1}$ and Dacheng Tao${}^{3}$	
}


\institute{Yunhe Wang, Kai Han, Hanting Chen, Chunjing Xu \at
              Noah's Ark Lab, Huawei Technologies. \\
           \and
           Mingqiang Huang, Wei Zhang\at
              Shenzhen Institutes of Advanced Technology, Chinese Academy of Sciences, Shenzhen 518055, China.\\
           \and
           Dacheng Tao\at
              School of Computer Science, Faculty of Engineering, the University of Sydney.               
}
\vspace{-5mm}
\date{Received: date / Accepted: date}

\maketitle

\begin{abstract}
 Convolutional neural networks (CNN) have been widely used for boosting the performance of many machine intelligence tasks. However, the CNN models are usually computationally intensive and energy consuming, since they are often designed with numerous multiply-operations and considerable parameters for the accuracy reason. Thus, it is difficult to directly apply them in the resource-constrained environments such as 'Internet of Things' (IoT) devices and smart phones. To reduce the computational complexity and energy burden, here we present a novel minimalist hardware architecture using adder convolutional neural network (AdderNet), in which the original convolution is replaced by adder kernel using only additions. To maximally excavate the potential energy consumption, we explore the low-bit quantization algorithm for AdderNet with shared-scaling-factor method, and we design both specific and general-purpose hardware accelerators for AdderNet. Experimental results show that the adder kernel with int8/int16 quantization also exhibits high performance, meanwhile consuming much less resources (theoretically $\mathrm{\sim}$81\% off). In addition, we deploy the quantized AdderNet on FPGA (Field Programmable Gate Array) platform. The whole AdderNet can practically achieve 16\% enhancement in speed, 67.6\%-71.4\% decrease in logic resource utilization and 47.85\%-77.9\% decrease in power consumption compared to CNN under the same circuit architecture. With a comprehensive comparison on the performance, power consumption, hardware resource consumption and network generalization capability, we conclude the AdderNet is able to surpass all the other competitors including the classical CNN, novel memristor-network, XNOR-Net and the shift-kernel based network, indicating its great potential in future high performance and energy-efficient artificial intelligence applications.


\end{abstract}

\section{Introduction}
\label{intro}
  Deep neural network has become ubiquitous in modern artificial intelligence tasks such as computer vision~\cite{imagenet,resnet,lecun2015deep} and natural language processing~\cite{transformer} due to its outstanding performance. In order to achieve higher accuracy, it has been a general trend to create deeper and more complicated neural network models, which are usually computationally intensive and energy consuming. However, in many real world applications, especially in the edge device platforms such as the sensors, drones, robotics and mobile phones, the complex recognition tasks have to be carried out in a resource-constrained condition with limited battery capacity, computation resource and chip area. Thus new deep learning models that can perform artificial intelligence algorithms with both high performance and low power consuming are strongly desired. 
  
  Many significant works have been proposed to address the model compression issue in the past several years~\cite{deepcompression,squeezenet,l1-pruning}. To directly decrease the computation amounts, network pruning~\cite{l1-pruning} is proposed to remove redundant weights thus compressing the original network. To construct efficient neural network with smaller model size, new convolution blocks and architectures are proposed. For example, SqueezeNet~\cite{squeezenet} introduces a bottleneck architecture and get the AlexNet-level accuracy by using only 2\% numbers of parameters. MobileNetV1~\cite{mobilenet} decomposes the conventional convolution filters into the point-wise and depth-wise convolution filters, thus much fewer ($\mathrm{\sim}$11\%) multiplication and accumulation (MAC) operations are involved in the new algorithm. MobileNetV2~\cite{mobilev2} uses the inverted residuals and linear bottlenecks to get further improvement compared to MobileNetV1. However, these efficient network architectures still rely on the multiplication-based convolution that is energy consuming. To make the convolution operation hardware-friendly and reduce the computation complexity of networks, low-bit model quantization methods are developed, in which the FIX16 and INT8 models are widely used due to its high accuracy and moderate model size~\cite{jacob2018quantization}. The binary/ternary weighted neural networks are proposed to further reduce the energy consumption~\cite{bnn,xnor}. However, these low-bit networks usually suffer from large degradation in their accuracy, especially for large datasets such as ImageNet. 
  
  Instead of the conventional multiplication-based convolution~\cite{lecun1998gradient}, several new deep learning computation manners including Deep-shift CNN~\cite{deepshift}, memristor CNN~\cite{yao2020fully,wang2019situ}, XNORnet~\cite{xnor} and mixed-precision CNN~\cite{chakraborty2020constructing} are proposed. The Deep-shift CNN replaces all multiplications with bit-wise shift, sign flipping operations and additions, but it still shows performance gap compared to the CNN baseline. The multi-bit version of Deep-shift exhibits much higher performance, but the energy consumption also increases. The memristor CNN integrates novel memristor~\cite{strukov2008missing} (a new kind of electronic device) arrays to implement the parallel MAC operations. By directly using Ohm's law for the multiplication and Kirchhoff's law for the accumulation, the in-memory computing structure of memristor network has greatly improved the energy-efficiency~\cite{wang2019situ,li2019long}. However, the network performance and its integration level are far from a practical artificial intelligence application. What is worse, the analogue circuit has to involve numbers of digital-to-analog and analog-to-digital converters (DAC/ADC), which will inevitably largely increase both the chip area and the power consumption~\cite{yao2017face}. Moreover, the conductance variation issue of the memristor is still highly desired to be conquered. The XNORnet and its improved version, namely the mixed-precision CNN, mainly take advantages of the the bit-level logic operation for the calculation, thus the energy consumption can be largely suppressed~\cite{chakraborty2020constructing}. However, their performance is also constrained by their low precision. In addition, the mixed-precision CNN allocates quantized weights with different bit-widths for different layers. It can be a meaningful compromise proposal to bridge the gap between the high performance and low power consumption, but it still contains massive multiplication operations and suffers the accuracy loss issue. 
  
  In this work, we present the adder-kernel convolutional neural network, where the convolutional kernel contains only adder operations~\cite{addernet}. AdderNet can fundamentally decrease both the logic resource consumption and the energy consumption, because adder operation is much lightweight than the multiplication operation~\cite{horowitz20141}. Besides, the AdderNet shows high generalization ability in versatile neural network architectures and performs well from pattern recognition to super-resolution reconstruction~\cite{xu2020kernel}. Experiments are conducted on several benchmark datasets and neural networks to verify the effectiveness of our approach. The performance of quantized AdderNet is very closed to or even exceeding to that of the CNN baselines with the same architecture. Finally, we thoroughly analyze the practical power and chip area consumption of AdderNet and CNN on FPGA. The whole AdderNet can achieve 67.6\%-71.4\% decrease in logic resource utilization (equivalent to the chip area) and 47.85\%-77.9\% decrease in energy consumption compared to the traditional CNN. In short, with a comprehensive discussion on the network performance, accuracy, quantization performance, power consumption, generalization capability, and the practical implementations on FPGA platform, the AdderNet is verified to surpass all the other competitors such as the novel memristor-based neural network and the shift-operation based neural network.

\section{Preliminaries} 
\label{sec:related}
\begin{figure*}
	\centering
	\includegraphics*[width=0.8\textwidth]{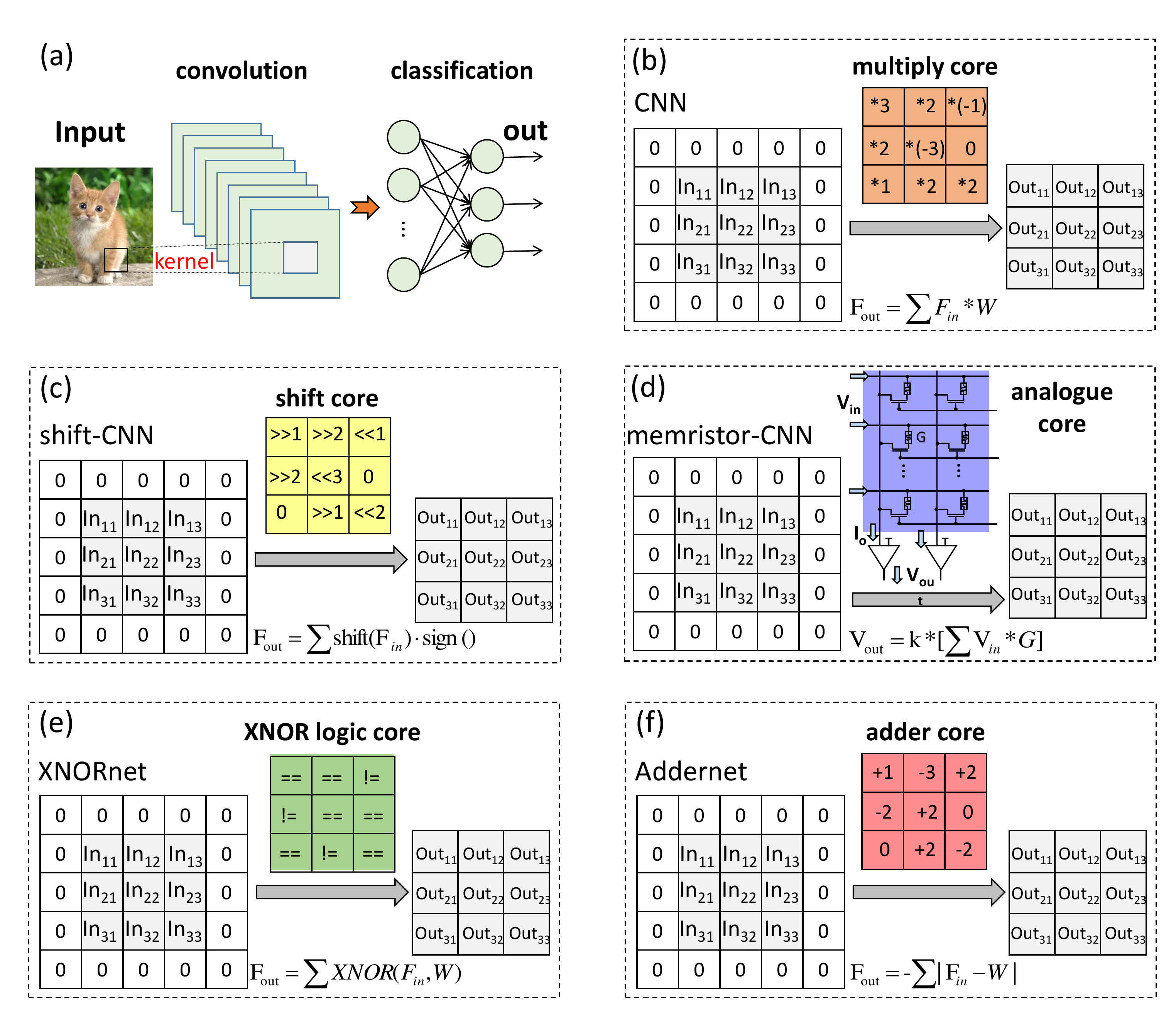} 
	\caption{Structure of Convolutional Neural Networks. \textbf{a})\textbf{ }The typical structure of convolutional neural network: input layer, a series of convolutional layers and pool layers, full-connection layer, and output layer. (\textbf{b}) The classical multiply core used in CNN. (\textbf{c}) The shift operation core used in Shift-CNN, in which both shift-operation and sign-operation are needed. If the data width of weight is larger than 1bit, adder-operations are also needed. (\textbf{d}) The analogue memristor core used in memristor-CNN. (\textbf{e}) The logic XNOR-operation core used in XNORnet (binary neural network). (\textbf{f}) The adder core used in AdderNet (this work).}
	\label{fig1}
\end{figure*}

	\subsection{Comparison of Different Convolutional Kernels}
	
	In deep convolutional neural networks (Figure~\ref{fig1}(a)), the convolutional layer is designed to detect local conjunctions of features using a series of convolution kernels (filters) by calculating the similarity between filters and inputs~\cite{lecun1998gradient,lecun2015deep}. Generally, there are 5 kinds of convolution kernels worth discussing, namely the multiplication-kernel in the traditional CNN (Figure~\ref{fig1}(b)), the shift-operation kernel (Figure~\ref{fig1}(c)), the analogue memristor kernel (Figure~\ref{fig1}(d)), the XNOR logic operation kernel (Figure~\ref{fig1}(e)), and the proposed adder-kernel (Figure~\ref{fig1}(f)). 
	
	For the representative CNN models, the vector multiplication (cross-correlation) is used to measure the similarity between input feature and convolution filter. Consider that the weight of filter is denoted as \textit{$W[K{}_{y}][K{}_{x}][CH{}_{in}][CH{}_{out}$]}, where \textit{K${}_{y}$}*\textit{K${}_{x}$} is kernel size, \textit{CH${}_{in}$} is input channel, \textit{CH${}_{out}$} is output channel, and the input feature is denoted as $F_{in}[H][W]$ $[CH{}_{in}]$, where $CH{}_{in\ }$is the number of feature channel\textit{, H} and \textit{W} are the height and width of the input feature, respectively. Then the output \textit{F${}_{out}$ }can be calculated by 
	\begin{equation} \label{GrindEQ__1_} 
	\begin{aligned}
	F_{{\rm out}} [h][w]&[ch_{out} ]=\sum _{{\rm k}_{{\rm y}} ={\rm 0}}^{K_{y} }\sum _{k_{x} =0}^{K_{x} }\sum _{ch_{in} =0}^{CH_{in} } S(F_{in} [h+k_{y} ]\\&[w+k_{x} ][ch_{in} ],W[k_{y} ][k_{x} ][ch_{in} ][ch_{out} ]),    
	\end{aligned}
	\end{equation} 
	where \textit{S(F${}_{in\ }$, W)} is a pre-defined similarity metric function. In CNN models, the metric function \textit{S} is multiplication, namely ${\rm S}({\rm F}_{{\rm in}} ,W)={\rm F}_{{\rm in}} \cdot W$. While in AdderNet, the L1-norm is used to measure the similarity between filters and inputs by calculating the absolute differences of \textit{F${}_{in}$ }and \textit{W, }namely ${\rm S}({\rm F}_{{\rm in}} ,W)={\rm -|F}_{{\rm in}} {\rm -}W{\rm |}$, in which the negative sign is to rectify the output result to be positive related with the L1-distance~\cite{addernet}.
	
	The Adder kernel proposed in this work is of great significance. On one hand, it has been widely believed that the custom large-scale convolutional neural networks have involved massive multiplication operations which consume much energy and computation resources. On the other hands, though there exists some other different kernels such as the shift-kernel~\cite{deepshift,shiftaddnet} and the memristor-kernel~\cite{yao2020fully}, they are exactly included in the classical multiplier-kernel. Particularly, the shift-operation is mathematically the same as multiplying by a positive (or negative) power of 2, and the memristor-kernel is only an analogue implementation of multiplier. The key point of this work is that we have experimentally shown that the CNN convolution kernel itself also contains redundant information and can be further simplified. In AdderNets, the L1-norm instead of cross-correlation in CNNs is taken to measure the similarity between features and filters, which is fundamentally different from the other kernels. 
	
	\begin{figure*}[htp]
		\centering
		\includegraphics*[width=0.9\textwidth]{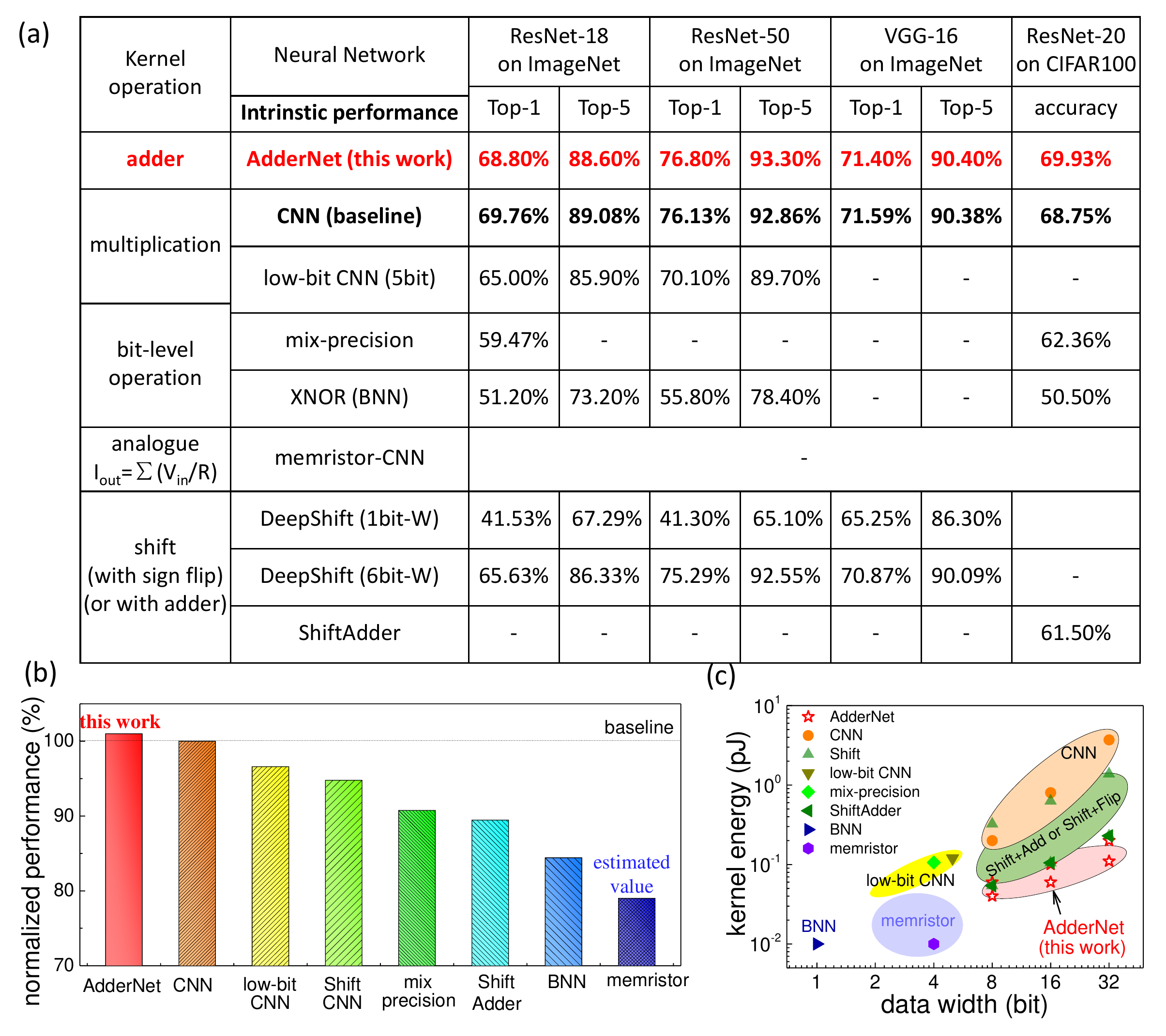} 
	\caption{Comparison of different kernels. (\textbf{a}) Recognition accuracy comparison of the different kernels, in which the CNN and our AdderNet exhibit best performance in all of the neural networks. Note, the current largest memristor-CNN model contains only 2 convolutional layers, thus there is no available data on the comparison table. (\textbf{b}) Normalized performance for different kernel methods. The data is summarized from table (a) and CNN is taken as the baseline. (\textbf{c}) Energy consumption of the different kernel operations}
	\label{fig2}
	\end{figure*}

	\subsection{AdderNet: High-Performance and Energy-Efficient Network }
	
	Performance is the most important metric for a neural network. Among the five kernels, the multiplication-kernel of CNN has been well explored and is considered to have the best performance~\cite{vgg,bn}. Impressively, the performance of our adder-kernel based AdderNet is comparable to or even better than that of CNN in many computer vision tasks. As shown in the performance comparison table of Figure~\ref{fig2}(a), the top-1 and top-5 accuracy of AdderNet ResNet-50 on ImageNet~\cite{imagenet} is 76.8\% and 93.3\%, respectively, and that of CNN is 76.13\% and 92.86\%. Besides, The accuracy of AdderNet ResNet-20 on CIFAR100 is 69.93\%, meanwhile that of CNN is only 68.75\%, where a 1.2\% accuracy improvement is achieved~\cite{xu2020kernel}. By contrast, the other kinds of convolution kernels (or their variations) show inferior performance. For example, the 6bit-weight Deepshift network and low-bit CNN usually exhibits $\mathrm{\sim}$1\% and 4\% accuracy degradation compared to CNN, respectively. And the 1bit-weight Deepshift~\cite{deepshift}, shiftaddnet~\cite{shiftaddnet}, mix-precision CNN~\cite{chakraborty2020constructing} and XNORnet~\cite{xnor} show even worse performance . 
	
	For the memristor network (or the other types of electronic synaptic devices such as the ferroelectric devices and phase change memory devives), though it has attracted great attentions in recent years, the performance and integration level are yet still far from a practical artificial intelligence application. To date, the state-of-the-art memristor network is just the demonstration of 2 layers CNN model~\cite{yao2020fully}. And its practical performance (recognition accuracy) on a small dataset (MNIST~\cite{lecun1998gradient}) is only about 79.76\%, which is much lower than all of the digital computation systems ($\mathrm{>}$95\%) mentioned above. What is worse, it could hardly get further improved due to the device conductance variation issue~\cite{huang2020global}.
	
	Figure~\ref{fig2}(b) shows the normalized network performance comparison for different convolution kernels, in which the data is extracted from Figure~\ref{fig2}(a). Obviously, the performance of all the other methods are lower than that of the original CNN baseline and our AdderNet. From high to low, the performance ranking list is AdderNet, CNN, 8-5 bit CNN, DeepShift, mix-precision CNN (depends on the precision), ShiftAdder, XNORnet(BNN) and memristor CNN, respectively.  
	
	Next, we would like to compare the circuit complexity and energy consumption for each kernel. The adder-kernel consists of one-Comparator-one-Adder (1C1A) or two Adders (2A) to calculate the absolute difference of weight and feature (Supplemental Information S1). The CNN multiplication-kernel contains one Multiplier. The 1bit shift-kernel consists of Serial-Shift-Register for the shift-operation and one Multiplexer as well as one N-bit*1-bit (N is the data width of feature) Multiplier for the sign function. Note, if the data width of weight (denoted as M-bit weight) in DeepShift is larger than 1, it also needs (M-1) numbers of N-bit adders and M groups of N-bit Serial-Shift-Registers. The memristor-kernel is made of two parallel 1-transistor-1-Memristor (1T1R) together with one differential circuit afterwards (Supplemental Information S2). The XNOR-kernel is the simplest one which contains only several AND/NAND logic gates (Supplemental Information S3).
	
	Supplemental Information S4 lists the power consumption for the different kernels~\cite{horowitz20141,shiftaddnet,thakre2015design}. It can be seen that the bit-level operation and analogue memristor cost the least consumption. The adder-operation and the shift-operation consume similar hardware resources and energy, and both can save a mass of consumption compared to the multiply-operation. For instance, the FP32 (float-point 32bit, a typical data type used in CPU or GPU) multiplication operation consumes 4.11$\mathrm{\times}$ and 1.84$\mathrm{\times}$ times of energy costs and logic area cost compared to the FP32 adder. And the FIX16 (fixed 16bit, a typical data type usually used in FPGA) multiplication operation consumes 15.7$\mathrm{\times}$ and 14.8$\mathrm{\times}$ times of energy and logic area cost more than that in FIX16 adder. The typical circuit area of each kernel can be checked in Supplemental Information S5.
	
	According to the above analysis, we have summarized the energy consumption in single kernel operation for the different networks in Figure~\ref{fig2}(c). The BNN and memristor network consume the least energy due to the bit-level operation and analogue circuits, respectively. However, these two achieves the worst network performance. Besides, the memristor networks have to involve numbers of digital-to-analog and analog-to-digital converters and will indeed largely increase the chip area as well as power consumption, which has not been considered in the kernel energy comparison. Our AdderNet exhibits extremely low power due to the low cost of adder-operation and comparator-operation. The low-bit (usually 4-8 bit) CNN and the shift-based (contains numbers of flip-operation or adder-operation) network exhibit a little bit higher computation resources but can still theoretically achieve about 50\%-90\% decrease in energy dissipation compared to CNN.

\begin{figure}[htp]
	\centering
	\includegraphics[width=0.48\textwidth]{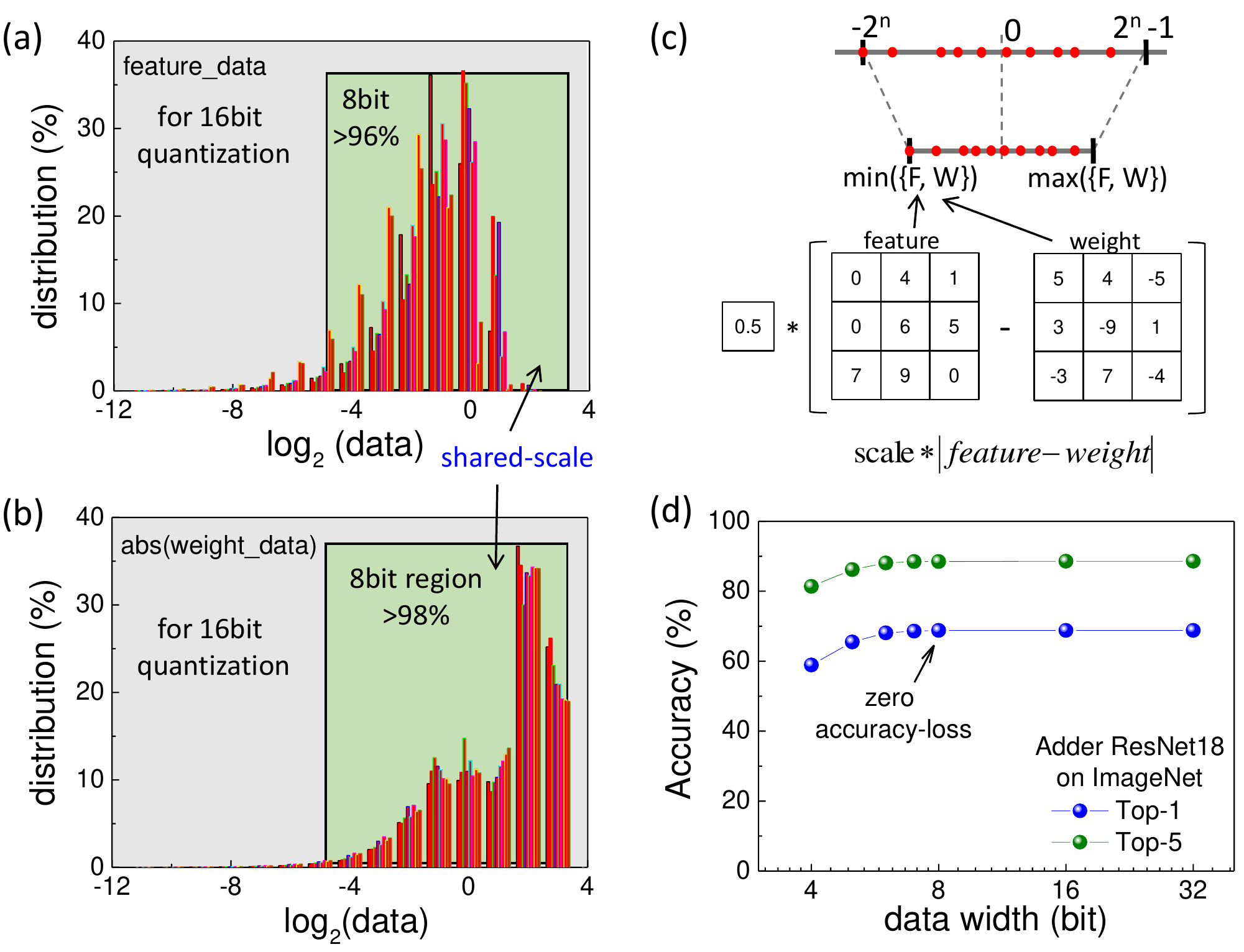} 
	\caption{Quantization with shared scaling factor. (\textbf{a}) and (b) Distribution of the input features and weights in different layers of AdderNet Resnet18. (c) Symmetric-mode of quantizations. (\textbf{d}) Quantization result of the AdderNet ResNet18. A Shared Scaling Factor for Feature and Weight quantization method is proposed, and the zero accuracy loss is achieved at 8bit quantization.}
	\label{fig3}
	\vspace{-2em}
\end{figure}

\section{Energy-efficient AdderNet on FPGA}

\subsection{Quantization with Shared Scaling Factor}

\begin{figure*}[htp]
	\centering
	\includegraphics[width=0.85\textwidth]{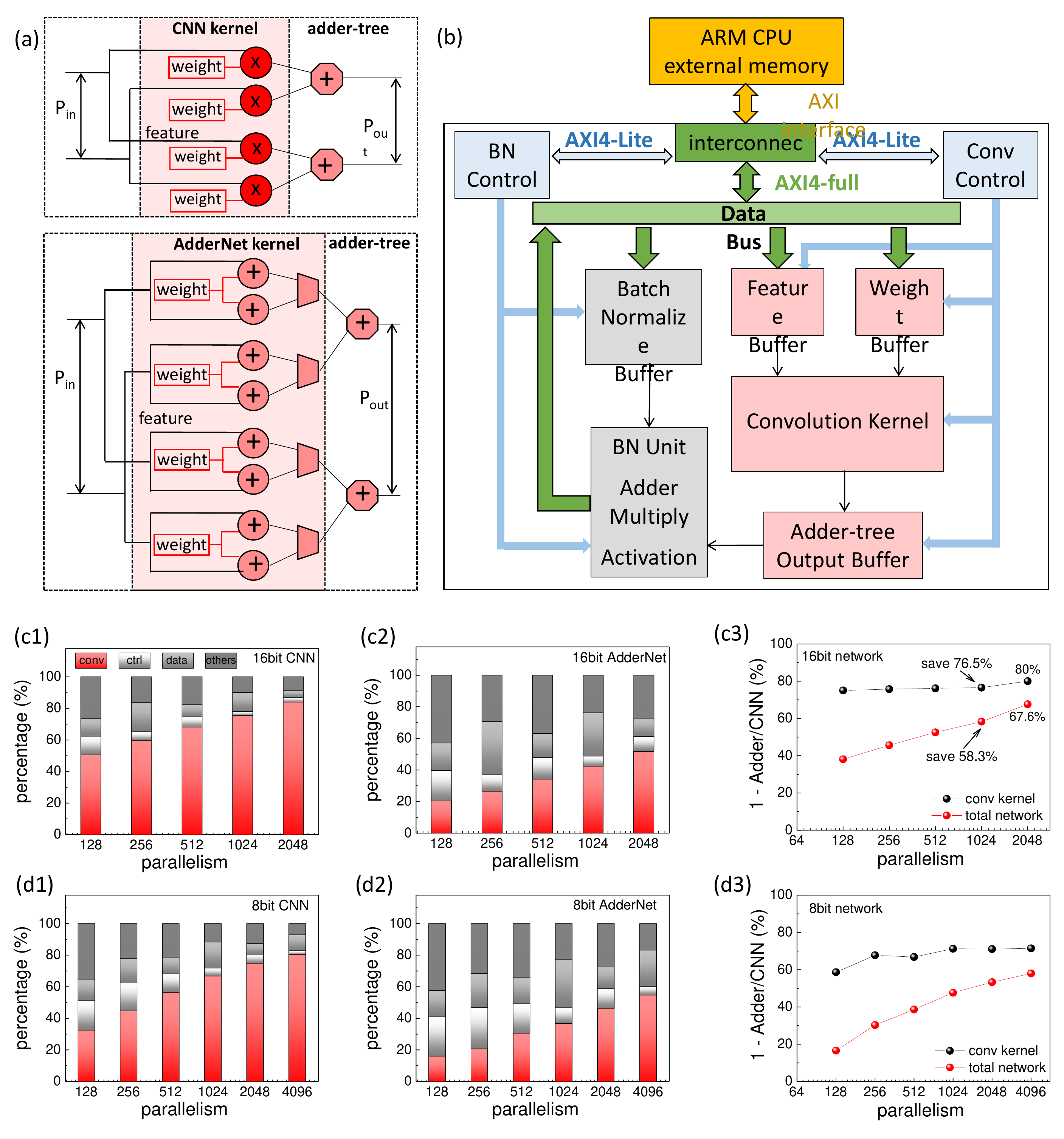} 
	\caption{Universal AdderNet accelerator. (\textbf{a}) Detailed structure of the FPGA accelerator that can be used for universal Convolutional Neural Networks. (\textbf{b}) Parallel computation on the convolution kernel. The AdderNet convolutional kernel mainly contains two adders and one multiplexer, and the CNN kernel contains one multiplier. Though the AdderNet kernel seems to be more complex, actually it is much lightweight than that of CNN. (\textbf{c1-c2}) The component of 16bit CNN and 16bit AdderNet at different parallelism levels synthesized in FPGA, respectively. To make a fair comparison, no DSP resources are used. It can be seen that the convolutional kernel occupies more and more logic resources during the increasing of parallelism. (\textbf{c3}) The convolutional part can save about 80\%-off in logic resource utilization,  and the total system is able to achieve 67.6\%-off. (\textbf{d1-d3}) Comparison results of AdderNet and CNN in 8bit network. The convolutional part and total network can save about 70\% and 58\% in logic resource utilization, respectively.}
	\label{fig4}
	\vspace{-1em}
\end{figure*}

Quantization is an important technique of model compression to represent weights and features with low-bit values~\cite{jacob2018quantization,bnn}. To achieve a balance between the high accuracy and low power consumption, 8bit or 16bit quantization scheme is often used. In CNN network, the input features and the weights are usually quantized with different scaling factors, thus the decimal points of features and weights are also different. And this would not introduce troubles in the low-bit MAC operation. However in AdderNet, if the features and weights were compressed with different scaling factors, the convolution parameters will have to be firstly point-aligned with shift operation before the adder operation, thus increasing the power and logic consumption in hardware. 

Figure~\ref{fig3}(a) and \ref{fig3}(b) show the distribution of input features and weights in AdderNet respectively, in which the different colors represent different convolution layers in ResNet18. It can be seen that the majority ($\mathrm{>}$90\%) of input feature ranges from 2${}^{-4}$ to 2${}^{2}$, and that of weight is from 2${}^{-2}$ to 2${}^{3}$. The relative small difference make it possible to quantize these two data sets together, namely the input features and weights can be quantized with the same scaling factor, which is different from that of CNN (separated scaling factor for feature and weight). Figure~\ref{fig3}(c) exhibits the typical quantization method. If the clip region was chosen to be 2${}^{-5}$ to 2${}^{3\ }$in 8bit quantization, 96\% of the feature information and 98\% of the weight information could be remained, thus guaranteeing the high performance. Besides, the shared scaling factor ensures that the adder convolution kernel in hardware can directly start the calculation without the need of point aligning, which is hardware friendly. 

The detailed performance of quantized AdderNet ResNet-18 has been shown in Figure~\ref{fig3}(d). The original Top-1 and Top-5 accuracies of the full-precision network are 68.8\% and 88.6\%, respectively. After the 8-bit quantization, the Top-1 and Top-5 accuracies remain 68.8\% and 88.5\% with near zero accuracy loss. Then, we further study the compression performance at even lower bit. As plotted in Figure~\ref{fig3}(d), the Top-1 and Top-5 accuracy can still reach 65.5\% Top-1 and 86.2\% Top-5 accuracies at 5-bit precision, respectively. While the 4-bit quantization exhibits relative large performance degradation, which is consistent with the above discussion with respect to the parameters distribution. The compression of ResNet-50 and its comparison to quantized CNN can be checked in Supplemental Information S6 and S7, respectively. 

\section{Hardware Design for Energy-efficient AdderNet}

The state-of-the-art hardware platforms for accelerating deep learning algorithms are field programmable gate array (FPGA), application specific integrated circuit (ASIC), Central processing unit (CPU), graphic processing unit (GPU), and digital signal processors (DSP). Among these approaches, FPGA based accelerators have attracted great attentions thanks to its good performance (much better than CPU, comparable to GPU), high energy efficiency, fast development round ($\mathrm{\sim}$several months, much faster than ASIC), and capability of reconfiguration . 

The structure of FPGA convolutional accelerator~\cite{chen2014diannao,chen2014dadiannao,zhang2016energy,zhang2015optimizing,ma2018optimizing} can be roughly divided into 4 parts: parallel kernel-operation core, data storage unit and Input/Output port, data path control module, and the others such as the Pooling and BN unit. Generally, the convolutional kernel core occupies most of the logic resources due to the Single Instruction Multiple Data (SIMD) of FPGA~\cite{zhang2015optimizing}. Therefore, the hardware simplification and optimization of convolutional kernel is of great significance. 

The numbers of input and output channels in convolutional network are usually to be the power of 2 (such as 64/128/256/512), thus making them suitable for the parallel computing. The typical examples are the Cambricon's DianNao~\cite{chen2014dadiannao,chen2014diannao}, Google's TPU~\cite{jouppi2017datacenter}, NVIDIA's open-source CNN accelerator, namely NVIDIA Deep Learning Accelerator (NVDLA)~\cite{nvidia}. In this work, we consider there are \textit{P${}_{in}$}${}_{\ }$input channels to be summed out in the adder tree for the \textit{P${}_{out}$} output channels (Figure~\ref{fig4}(a)). Suppose the data width in network is \textit{DW}, then the consumption of AdderNet in parallel convolutional kernel is "\textit{P${}_{out}$*(P${}_{in}$*DW*2)}", where the number of "\textit{2}" comes from the fact that there are two adders in each AdderNet kernel. The consumption of AdderNet in adder-tree is "\textit{P${}_{out}$*([DW+log${}_{2}$(P${}_{in}$)]*(P${}_{in}$-1))}", in which the "\textit{(P${}_{in}$-1)}" and "\textit{[DW+log${}_{2}$(P${}_{in}$)]}" represents the numbers of adder and the data width in adder tree, respectively. Therefore the theoretical logic resource consumption (and also energy consumption) of AdderNet is 
\begin{equation} \label{GrindEQ__2_} 
{\rm P}_{{\rm out}} {\rm*}\{{\rm P}_{{\rm in}} *DW*2+[DW+\log _{2}(P_{in})]*(P_{in} -1)\}. 
\end{equation} 

The consumption of CNN in parallel convolutional kernel can be calculated by "\textit{P${}_{out}$*(P${}_{in}$*DW*DW)}", where the second "\textit{DW}" represents the theoretical value of multiply to adder ratio. The consumption in adder-tree is "$P{}_{out}*([2*DW+log{}_{2}(P{}_{in})-1]*(P{}_{in}-1))$", in which "\textit{(P${}_{in}$-1)}" represents the numbers of adder, "\textit{2*DW}" is the data width after multiplication kernel, and "\textit{[2*DW+log${}_{2}$(P${}_{in}$)-1]}" is the data width in adder tree. Thus the theoretical resource consumption of CNN is 
\begin{equation} \label{GrindEQ__3_} 
\begin{aligned}
&{\rm P}_{{\rm out}} *\{{\rm \; P}_{{\rm in}} *DW*DW+\\& [DW*2+\log _{2}^{} (P_{in} )-1]*(P_{in} -1)\}.
\end{aligned}
\end{equation} 
If the \textit{DW} is fixed at 16 and $P{}_{in}$ is designed to be 64, the AdderNet will theoretically get 81.6\%-off in logic resource utilization (and also power consumption) compared to CNN. However, the practical count of energy-efficiency improvement for AdderNet is still a challenge. This mainly comes from the Data Input/Output overhead, where the data move from the outside main Memory (Dynamic Random Access Memory, DRAM) to the computation part will cause an enormous amount of energy consumption. To verify the accurate energy and logic resource consumption, we have designed two kinds of hardware implementations as following. 

Figure~\ref{fig4}(b) shows the overview design of our general purpose accelerator implemented on Xilinx Zynq UltraScale+ MPSoC series Board which integrates the Processing System (PS, namely the ARM Cortex-series CPU) and the Programmable Logic (PL, namely the programmable FPGA) together. The convolution accelerator is composed of external memory, on/off-chip interconnect, on-chip buffer, and the parallel computation procession element. The data transportation between the PS and PL are controlled by the Advanced eXtensible Interface (AXI) interconnection protocol, in which the AXI-lite protocol is used for the parameter configuration and AXI-full protocol is used for the weight/feature transposition. 

\begin{figure*}[htp]
	\centering
	\includegraphics[width=0.9\textwidth]{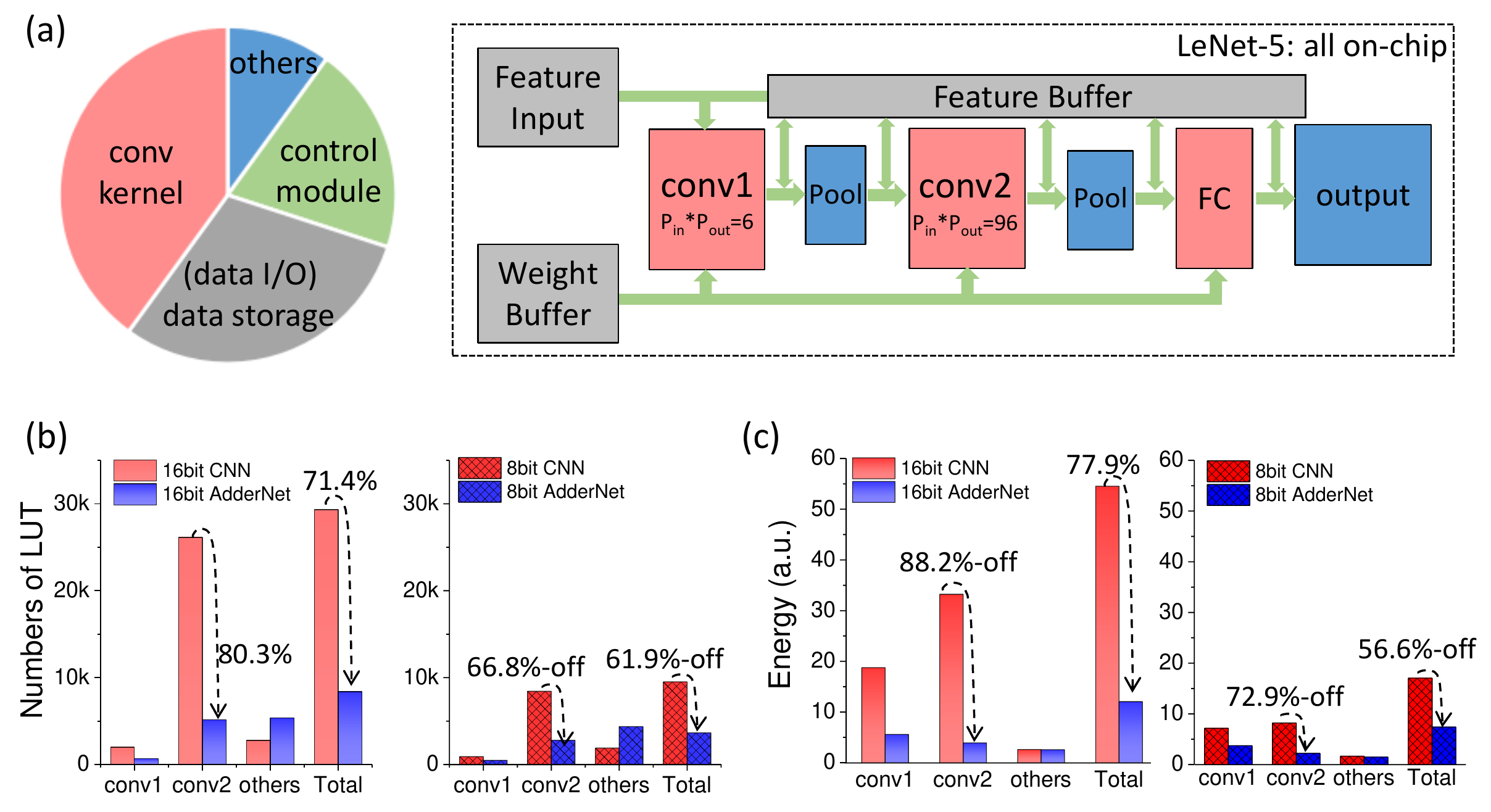} 
	\caption{FPGA AdderNet without off-chip data excess. (\textbf{a}) Left: consists of a typical FPGA Convolutional Neural Networks accelerator: the multi-channel kernel-operation core, data storage (including both data buffer and Data Input/Output), data path control module, and the other modules such as the Pooling and BN unit. Righ: Structure of the FPGA based CNN accelerator for LeNet-5, in which no Data Input/Output overhead is needed and all the calculations and data storage will be implemented on board. (\textbf{b}) and (\textbf{c}) Comparison of AdderNet and CNN on the logic resource utilization and energy consumption, respectively.}
	\label{fig5}
\end{figure*}

Figure~\ref{fig4}(c) and ~\ref{fig4}(d) shows the synthesized result of the 16bit and 8bit convolution path in FPGA, respectively. To make a fair comparison, no DSP resources are used here and all of the modules in the accelerator are constructed by using the LUT (look-up-table in FPGA) resources. Figure~\ref{fig4}(c1) draws the component (including the convolution kernel, data storage/buffer, data path control module, and others) distribution of logic resources at different parallelism levels. It can be seen that the convolution kernel occupies more and more logic resources during the increasing of parallelism. For example, when the parallelism is 128, the computation unit occupies 50.48\% in the whole system. When the parallelism is 2048, the computation unit will dominate and take up about 83.9\% in the total logic resources. Similar result can be achieved in the 8 bit AdderNet as shown in Figure~\ref{fig4}(c2). 

The more important thing is how much of the logic resource that AdderNet can save compared to CNN. Figure~\ref{fig4}(c3) plots the result, in which the red dots represent the convolution kernel and the black dot represent the total accelerator. It is clear to see that the total logic resource that AdderNet can save compared to CNN increases with the parallelism level. When parallelism = 2048, the AdderNet can get 67.6\%-off for the total system, and 80\%-off in the convolutional part, which is consistent with the theoretical value according to equation \eqref{GrindEQ__2_} and \eqref{GrindEQ__3_}. Figure~\ref{fig4}(d) shows the similar comparison results of AdderNet and CNN in 8bit network. The 8bit AdderNet can save about 70\%-off in convolution computation and 58\%-off in the total system, and the value can be further improved if the parallelism goes even larger. 

Then we will tend to the practical comparison of AdderNet and CNN on board (Xilinx Zynq UltraScale+ MPSoC ZCU104). Due to the limited logic resources in ZCU104, the parallelism of CNN is restrained to be 1024, and so is the AdderNet. The first advantage of AdderNet is the capability for operating at higher frequency with higher performance. Because the multiplier (in combination logic circuit design) owns much higher logic gate delay (T${}_{comb}$) compares to adder, it is difficult for CNN to get the positive setup-time and hold-time in static timing analysis especially in high frequency. Here in our design, the highest operation frequency of CNN is 214 MHz, and that of AdderNet is 250 MHz. For the running of ResNet18, the CNN exhibits 424 GOPs (Giga operations per second) for the convolution calculation and 307 GOPs for the whole network. While the AdderNet is able to achieve as high as 495 GOPs for the convolution and 358.6 GOPs for the whole network, which is among the best results in the similar hardware platform (Supplemental Information S8).

Besides, the operating power of adder convolution is much lower than that of CNN. We have measured the practical power consumption of AdderNet and CNN at 214 MHz during convolution, in which all of the activities are counted, such as the off-chip data excess, data input/output transportation, control module, and the convolution. After subtracting the ZCU104 embedded system operation power (can be treated as the baseline or background noise, $\mathrm{\sim}$14 W), the intrinsic power of AdderNet convolution is 1.34 W, while that of CNN is 2.57 W, which means AdderNet is able to get 47.85\%-off in the practical energy consumption.

The above deviation from the theoretical value mainly comes from the data transportation between on-chip computation part and the off-chip memory. To avoid this, we deploy a small-scale network (i.e., LeNet-5) on Zynq-7020 (with Xilinx XC7Z020 chip), in which no Data Input/Output overhead is needed, and all of the weight parameters and intermediary computation results are stored on board. As shown in Figure~\ref{fig5}(a), the LeNet-5 contains 2 convolutional layers, where the first one is with 1 input channel and 6 output channels, and the second one is with 6 input channels and 16 output channels. To accelerate the convolution operation, we use the parallel computation technology in both input and output channels. Therefore, there are 6 parallel kernel operators for the first convolutional layer and 96 for the second layer. 

Figure~\ref{fig5}(b) and \ref{fig5}(c) show the comparison results of CNN and AdderNet. For the 16bit network, our AdderNet gets 70.3\%-off, 80.32\%-off and 71.4\%-off compared to CNN in number of LUTs (equivalent to the chip area) in the conv-layer1, conv-layer2 and total network, respectively. Besides, the drop of energy consumption in AdderNet can reach 70.22\% -off, 88.29\%-off, and 77.91\%-off, respectively. The result is quite close to the estimated value $\mathrm{\sim}$81\%. For the 8bit network, the AdderNet gets 46.76\%-off, 66.86\%-off and 61.63\%-off than CNN in logic resource utilization, and 48.33\%-off, 72.96\%-off, 56.57\%-off in energy consumption, which is also close to the estimated value. 

\section{Implementation Details}
\paragraph{Optimization of AdderNet.} 
The back-propagation algorithm is utilized for calculating the gradients of AdderNet~\cite{addernet} just as the conventional neural networks. The loss function for image recognition task is cross-entropy loss. To improve the performance of AdderNet, we also apply the distillation loss~\cite{xu2020kernel} on AdderNet by using CNN as teacher networks (Supplemental Information S9).

\paragraph{Datasets.} 
We have verified the performance of AdderNet on various computer vision datasets including CIFAR100 and ImageNet. CIFAR100 dataset~\cite{cifar} consists of 50,000 32x32 colour training images and 10,000 test images from 100 classes. ImageNet ILSVRC2012~\cite{imagenet} is a large-scale image classification dataset which is composed of about 1.2 million training images and 50,000 validation images belonging to 1000 categories. The commonly used data augmentation strategy including random crop and random flip is adopted for image pre-processing~\cite{resnet}.

\paragraph{Training details.} 
All the networks are trained on NVIDIA Tesla V100 GPUs using PyTorch. For CIFAR100, the models are trained for 400 epochs with a batch size of 256. The learning rate starts from 0.1 and decays using cosine scheduler. The weight decay is set as 0.0005. For ImageNet, the models are trained for 150 epochs with a batch size of 256. The learning rate starts from 0.1 and decays using cosine scheduler. The weight decay is set as 0.0001.

\paragraph{Hardware information.} 
The accelerator is designed and simulated with Vivado (v2018.1). The hardware platform for the large-scale network is the Zynq UltraScale + MPSoC ZCU104 board which has a Xilinx FPGA chip of XCZU7EV-2FFVC1156 MPSoC, and the small-scale network is deployed on ZYNQ7020 board with Xilinx XC7Z020 chip. The hardware implementation runs on PYNQ (Python Productivity for ZYNQ) zcu104\_v2.5 environment.

\section{Conclusion}

This paper studies the low-bit quantization algorithm for adder neural network (AdderNet) and the hardware design for achieving the state-of-the-art performance in terms of both accuracy and hardware overhead. Specifically, the AdderNet exhibits high performance and low power advantages over its counterparts such as the conventional multiply-kernel CNN, the novel memristor-kernel CNN, and the shift-kernel CNN. Besides, we propose the shared-scaling-factor quantization method for AdderNet, which is not only hardware friendly but also showing guaranteed performance with nearly zero accuracy loss down to 6bit network. Finally, we design both specific and general-purpose convolution accelerators for implementing neural networks on FPGA. In practice, the adder kernel can theoretically obtain an about 81\%-off in resource consumption compared to multiplication kernel (i.e., the conventional CNN). Experimental results illustrate that  by applying the AdderNet, we can achieve  about 1.16$\mathrm{\times}$ speed-up ratio, 67.6\%-71.4\% decrease in logic resource utilization (chip area), 47.85\%-77.9\% decrease in power consumption compared to CNN with exactly the same circuits design, respectively.

\section*{Acknowledgments}
This work was supported by Shenzhen Science and Technology Innovation Committee JCYJ 20200109115210307.

\clearpage
\section*{Supplemental Information}
\subsection*{\textbf{S1: Details of adder convolution kernel}}
To calculate the absolute difference of weight and feature, the adder-kernel can be designed by using one-Comparator-one-Adder (1C1A) scheme or two Adders (2A) scheme (Figure~\ref{fig-s1}). The 1C1A scheme is to compare the input of $A_{in}$ and $B_{in}$ firstly, then get the subtraction value of the larger one minus the smaller one. The advantage of such design is the less resource consumption of digital comparator. However, it suffers from the relative larger gate-delay, which may decrease the circuit speed. The 2A scheme is able to operate at higher frequency due to the parallel two adders, but the adder is more complex than that of comparator, thus the circuit area will be higher. In this work, we choose the 2A scheme for a higher performance.

\subsection*{\textbf{S2: Memristor and memristor network}}
The reason that memristor can be used in neural network is that its conductance can be modulated by electrical pulse. Figure~\ref{fig-s2}) shows one typical memristor and its performance. The conductance can be tuned from $G_{min}$ to $G_{max}$ in several discrete levels, which is similar to the weight parameters in network. However, the conductance level in memristor is usually limited to be only 4-6 bit, and the conductance variation issue is avoidless.

The memristor network is usually designed with one-Transistor-one-Memristor (1T1R) structure, in which the transistor is to individually control the memristor by using the write-line (WL) and select-line (SL). For the conductance in memristor is always positive, while the weight parameter in network can be either positive or negative, the differential circuit of 1T1R-1T1R is used (green region in figure S2b). The multiplication accumulation (MAC) can be realized by directly using Ohm's law and Kirchhoff's law in memristor array. However, it needs great numbers of digital-to-analog and analog-to-digital converters (DAC/ADC) in the periphery controller module, which will inevitably largely increase both the chip area and the power consumption.

\subsection*{\textbf{S3: Binary neural network and Binary memristor network}}
The XNOR-kernel based binary neural network (BNN) consumes the lest logic resources and energy, because the XNOR contains only several AND/NAND logic gates, which is much lightweight than all of the other kernels.

The combination of binary neural network and memristor network is the novel Binary memristor network~\cite{huang2020global}. It is light-weighted, highly robust and can work well on challenging visual tasks. However, the device may suffer the stunk of Set/Reset state (Figure~\ref{fig-s3}(d)), and it also needs DAC/ADC in the periphery circuits.

\begin{figure}[htp]
	\centering
	\includegraphics[width=0.8\linewidth]{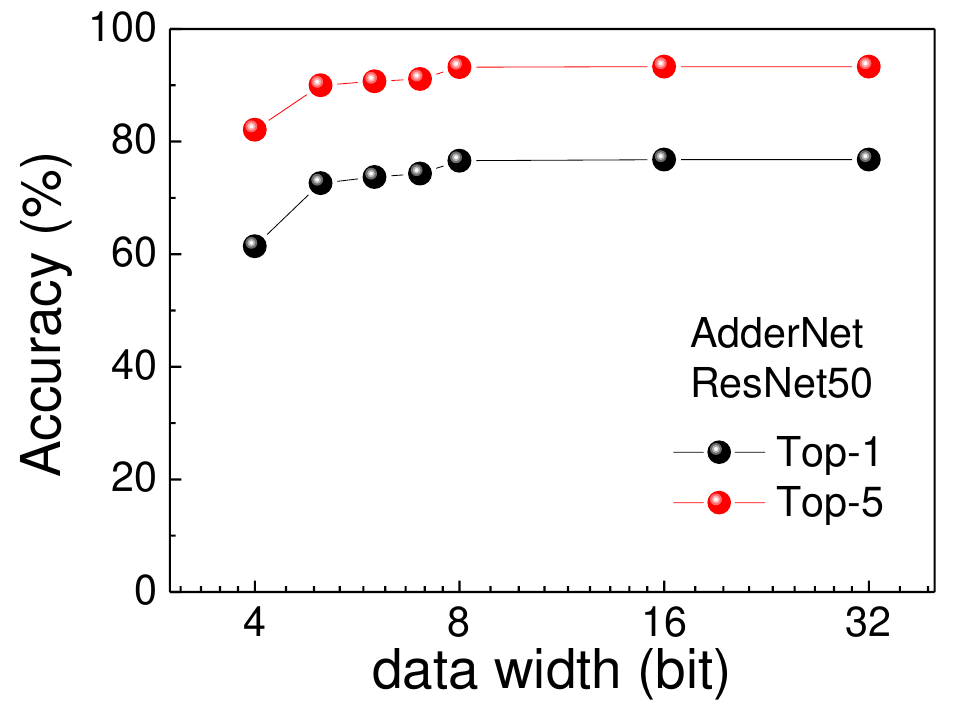} 
	\caption{The original Top-1 and Top-5 accuracies of the full-precision ResNet-50 network are 76.8\% and 93.3\%, respectively. After the 8-bit quantization, the Top-1 and Top-5 accuracies can still reach 76.6\% and 93.2\%, respectively.}
	\vspace{-1em}
	\label{fig-s6}
\end{figure}

\begin{figure}[htp]
	\centering
	\includegraphics[width=0.8\linewidth]{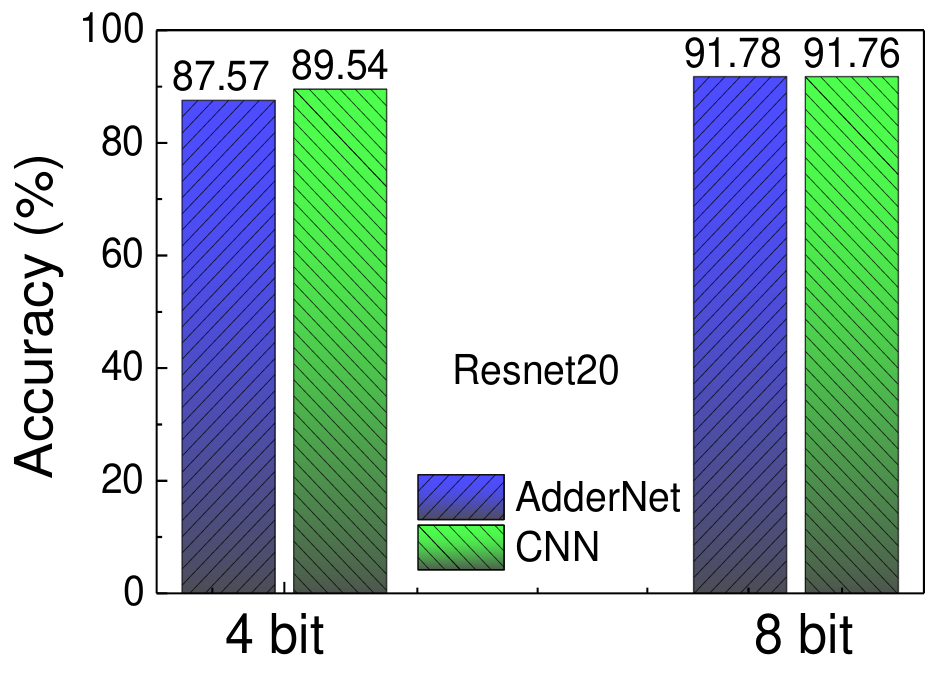} 
	\caption{The accuracy of CNN-ResNet20 at 8bit and 4bit is 91.76\% and 89.54\%, respectively. While that of AdderNet-ResNet20 is 91.78\% and 87.57\%, respectively. It can be seen that the AdderNet shows similar performance with CNN at higher bit precision. However, if the data width decreased down to 4bit, the AdderNet accuracy will degrade largely. This is because the Shared-Scale-Factor in AdderNet quantization may loss more information.}
	\label{fig-s7}
	\vspace{-2em}
\end{figure}

\subsection*{\textbf{S4: Detailed comparison of energy consumption in different convolution kernels}}
We compare energy consumption in different convolution kernels including AdderNet, CNN, DeepShift~\cite{deepshift}, XNOR-net~\cite{xnor} and memristor~\cite{yao2020fully}. The results are shown in Table~\ref{fig-s4}.

\begin{figure*}[htp]
	\centering
	\includegraphics[width=0.85\textwidth]{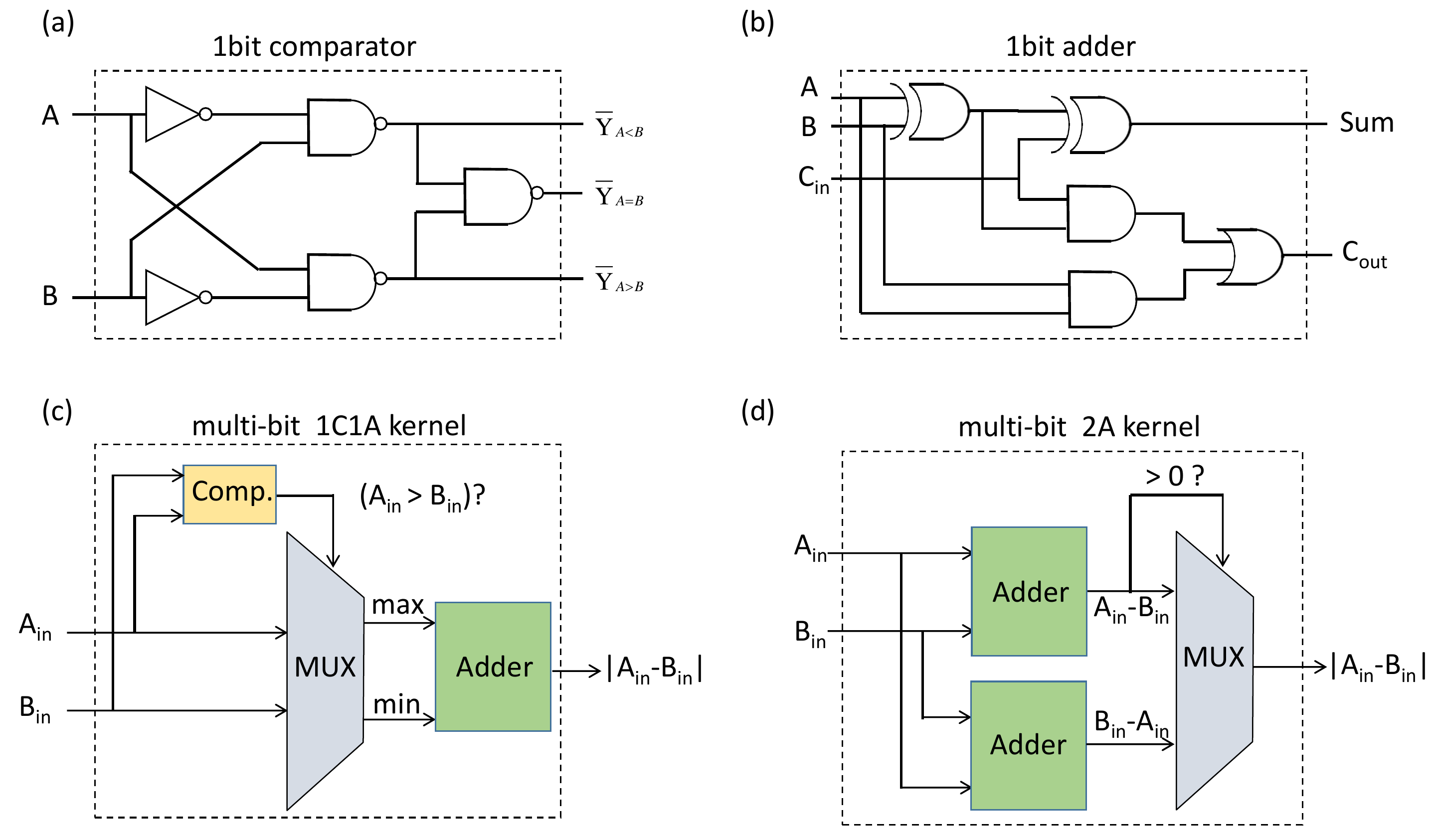} 
	\caption{(a) and (b) Logic circuit of 1-bit comparator and 1-bit Adder. The adder is more complex than that of comparator. (c) and (d) The typical design of adder convolution kernel in 1C1A and 2A, respectively. The multiplex (MUX) is constructed by several AND gates, and is much lightweight than other logic parts.}
	\label{fig-s1}
\end{figure*}

\begin{figure*}[htp]
	\centering
	\includegraphics[width=0.9\textwidth]{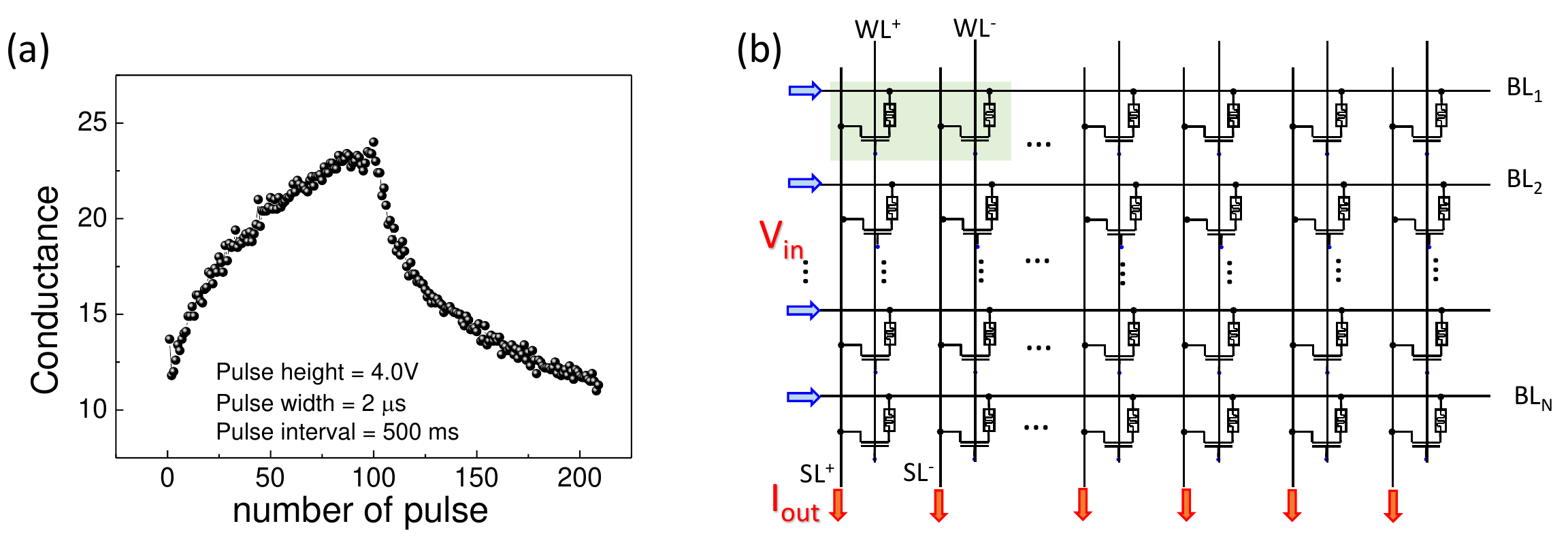} 
	\caption{(a) Performance of a typical memristor. (b) Structure of memristor network.}
	\label{fig-s2}
\end{figure*}

\subsection*{\textbf{S5: Detailed comparison of logic circuit area in different convolution kernels}}
We compare energy consumption in different convolution kernels including AdderNet, CNN, DeepShift~\cite{deepshift}, XNOR-net~\cite{xnor} and memristor~\cite{yao2020fully}. The results are shown in Figure~\ref{fig-s5}.

\subsection*{\textbf{S6: Quantization of ResNet-50}}
The quantization results with the proposed shared-scale quantization method are shown in Figure~\ref{fig-s6}.

\subsection*{\textbf{S7: Performance comparison of AdderNet and CNN after Quantization}}
We compare the quantization resultsof AdderNet and CNN as shown in Figure~\ref{fig-s7}.

\subsection*{\textbf{S8: Performance comparison of different FPGA neural network}}
We compare energy consumption in different FPGA neural networks including models in \cite{rahman2016efficient}, \cite{li2016high}, \cite{aydonat2017opencl}, \cite{guo2017angel}, \cite{zhang2016energy}, \cite{wei2017automated}, \cite{guan2017fp}. The results are shown in Figure~\ref{fig-s8}.

\begin{figure*}[htp]
	\centering
	\includegraphics[width=0.8\textwidth]{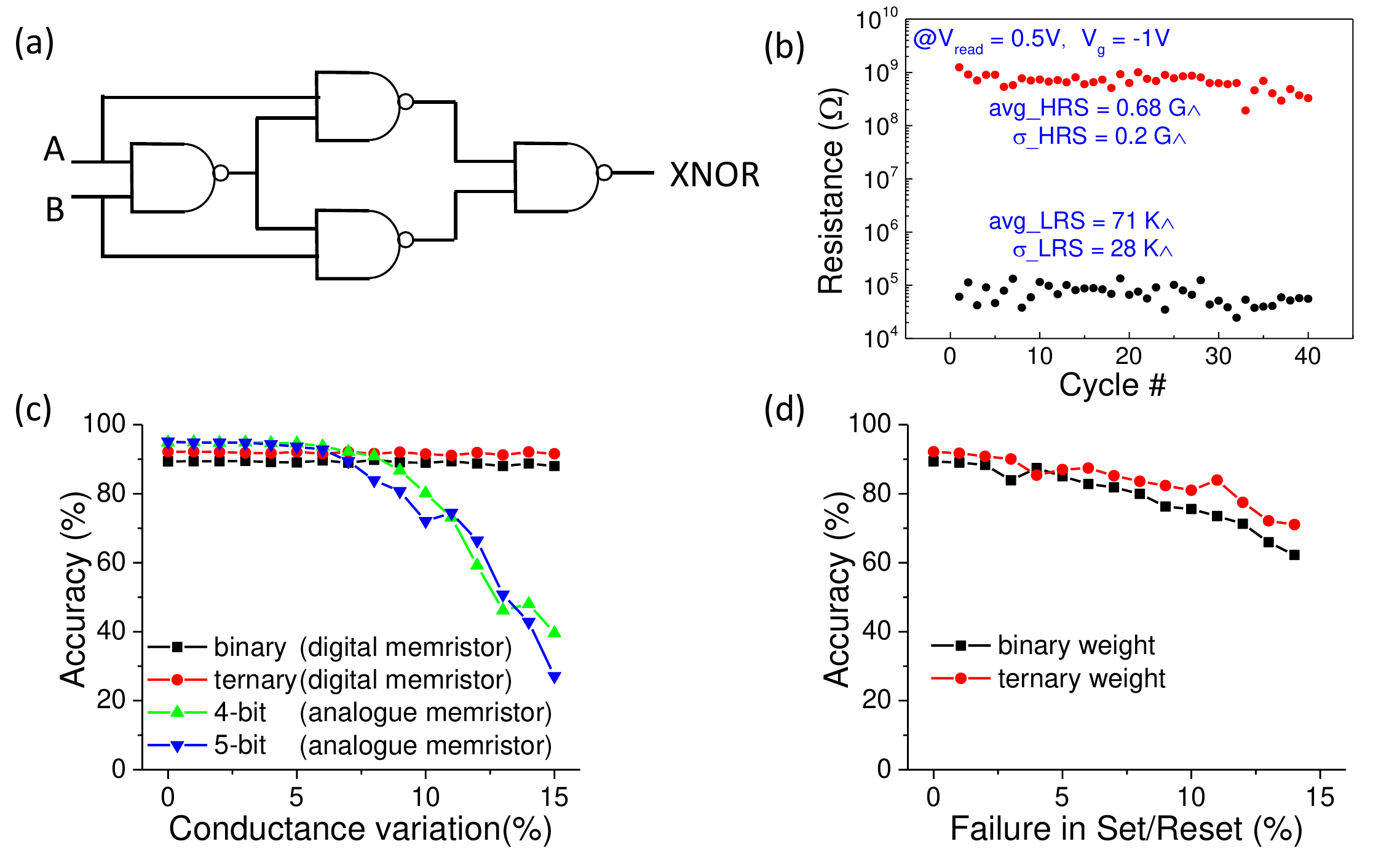} 
	\caption{(a) The XNOR gate used in binary neural network. (b) to (d) Performance of the Binary memristor network.}
	\label{fig-s3}
\end{figure*}

\begin{figure*}[htp]
	\centering
	\includegraphics[width=0.75\textwidth]{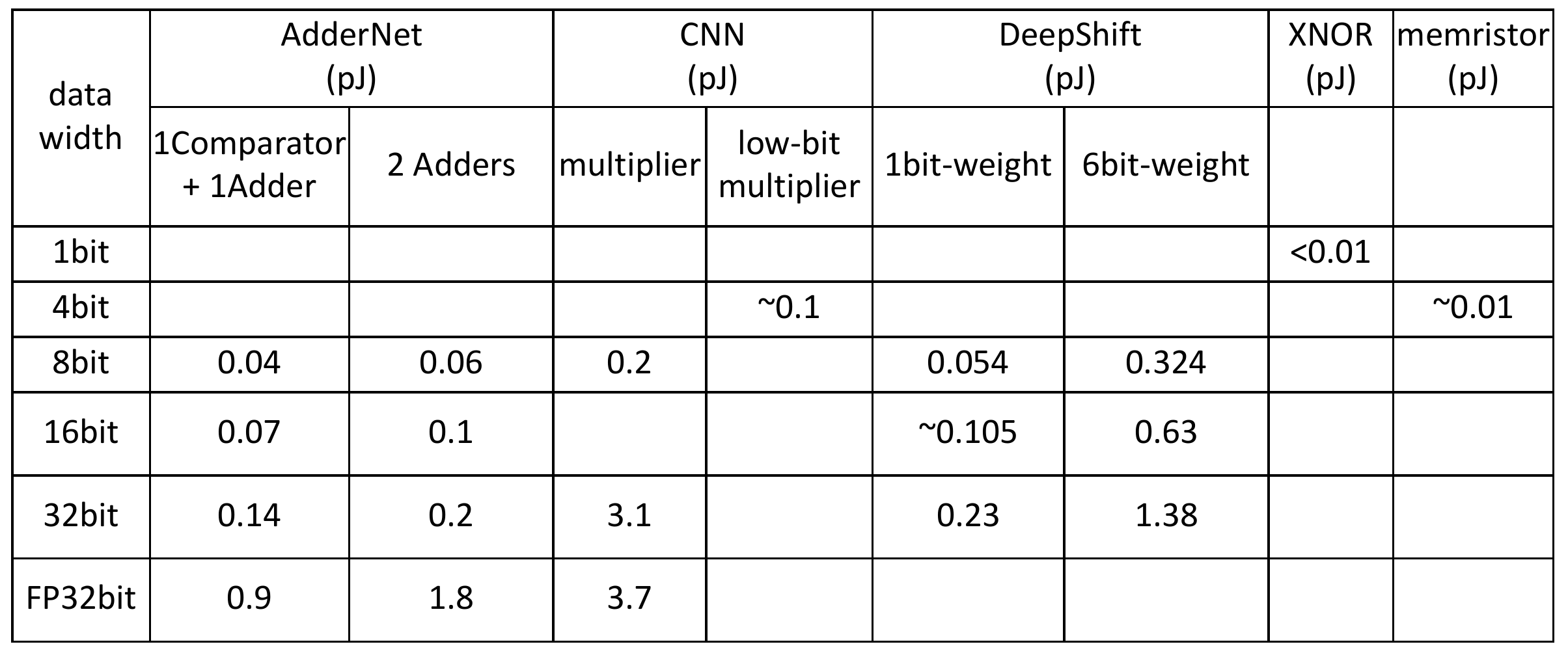} 
	\caption{Detailed comparison of energy consumption in different convolution kernel~\cite{horowitz20141,thakre2015design,shiftaddnet}.}
	\label{fig-s4}
\end{figure*}

\begin{figure*}[htp]
	\centering
	\includegraphics[width=0.75\textwidth]{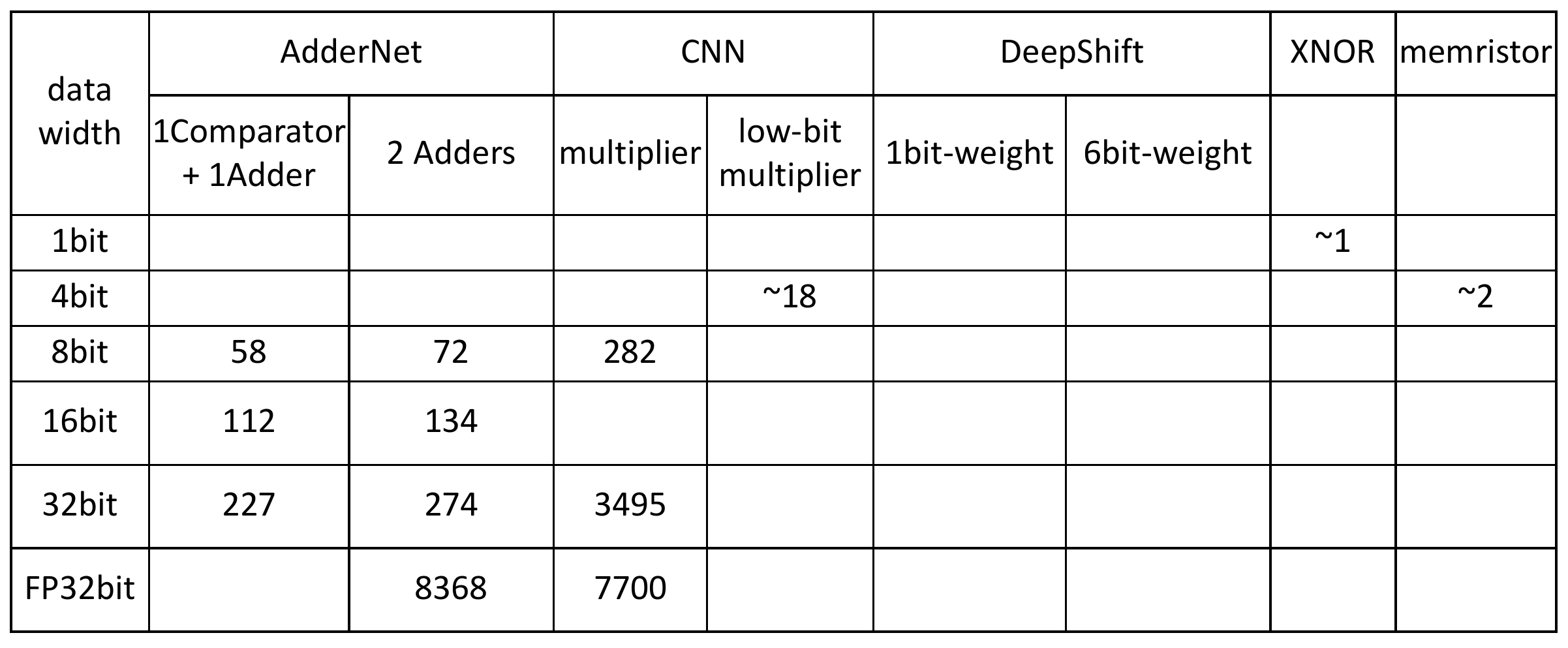} 
	\caption{Detailed comparison of logic circuit area in different convolution kernels~\cite{thakre2015design}.}
	\label{fig-s5}
\end{figure*}

\begin{figure*}[htp]
	\centering
	\includegraphics[width=0.9\linewidth]{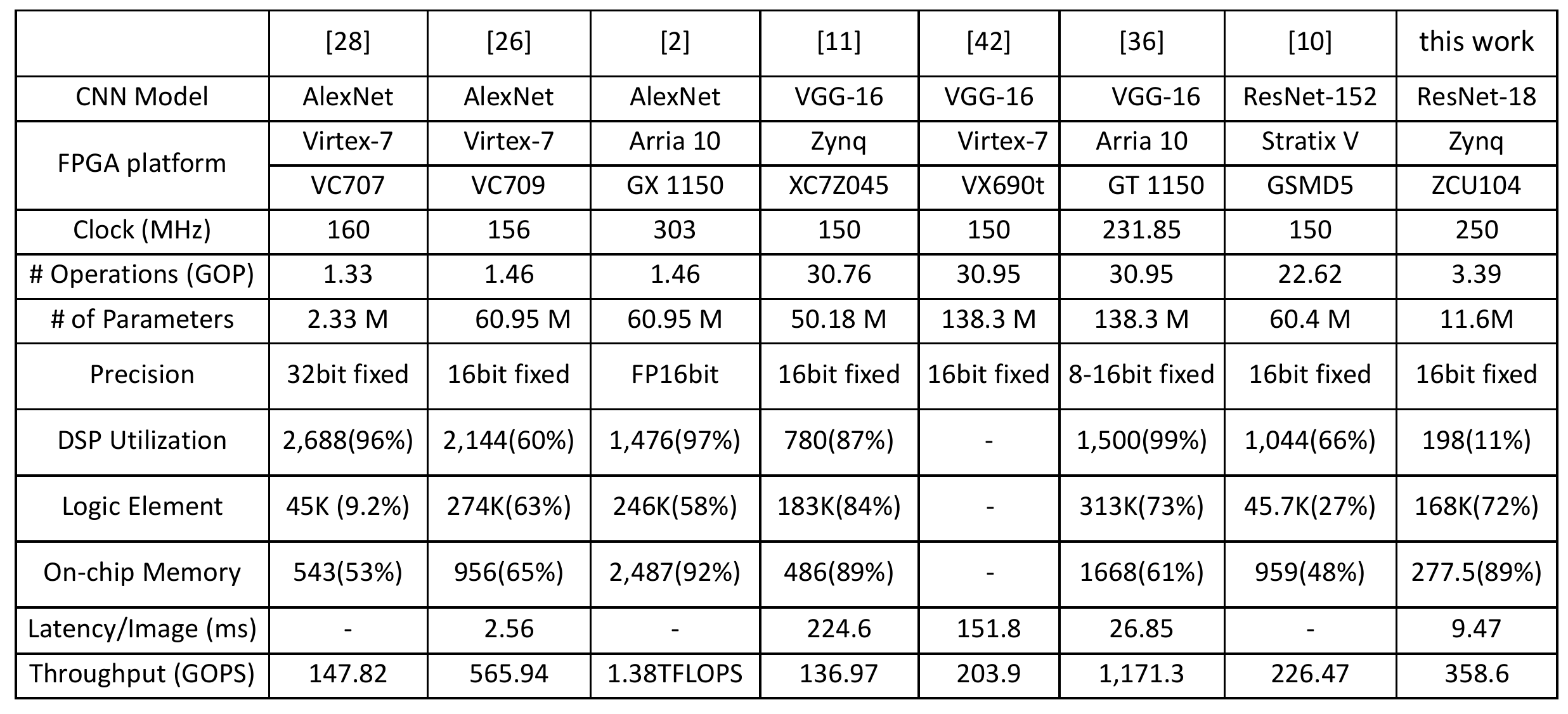} 
	\caption{Performance comparison of different FPGA neural network.}
	\label{fig-s8}
\end{figure*}

\subsection*{\textbf{S9: Accuracy and loss of AdderNet in training}}
\begin{figure}[htp]
	\centering
	\includegraphics[width=0.9\linewidth]{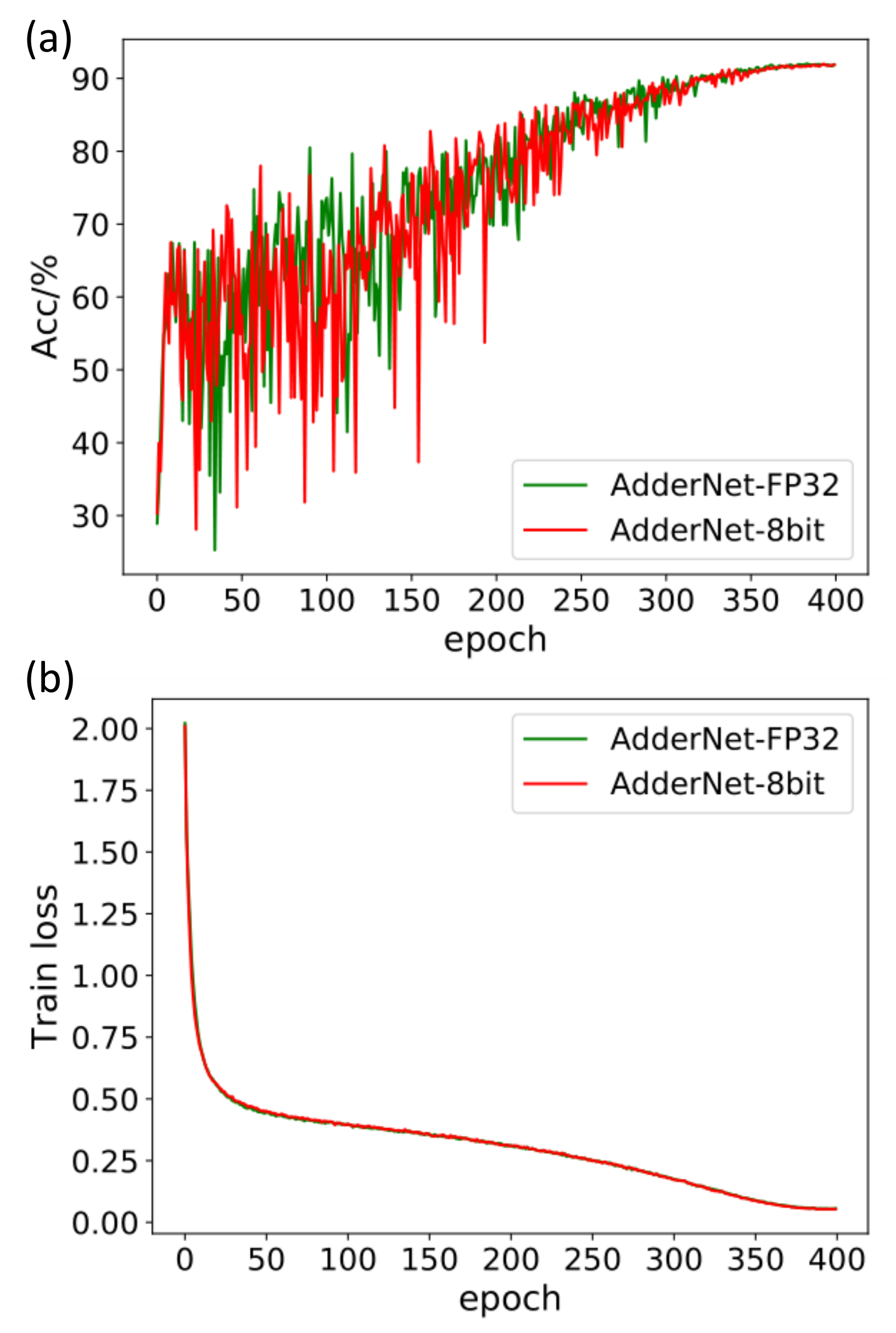} 
	\caption{Accuracy and loss of AdderNet in training.}
	\label{fig-s9}
\end{figure}

We plot the accuracy and loss curves for better understanding the training of AdderNet (Figure~\ref{fig-s9}).

{\small
	\bibliographystyle{plain}
	\bibliography{mybib}
}
\end{document}